\pdfoutput=1

\documentclass[11pt]{article}

\usepackage{naacl2021}

\usepackage{times}
\usepackage{latexsym}

\usepackage[T1]{fontenc}

\usepackage[utf8]{inputenc}

\usepackage{microtype}

\usepackage{amsmath,amssymb}
\usepackage{mathtools}
\usepackage{subcaption}
\newcommand\sS{\mathbf{s}}

\title{R2D2: Relational Text Decoding with Transformers}

\author{Aryan Arbabi \\
  University of Toronto\\
  Vector Institute\\
  \texttt{arbabi@cs.toronto.edu} \\\And
  Mingqiu Wang \and Laurent El Shafey\\ 
\and {\bf Nan Du} \and \bf{Izhak Shafran} \\
    Google Research\\
  \texttt{\{mingqiuwang,dunan,shafey,izhak\}}\\
  @google.com\\}

\begin{document}
\maketitle
\begin{abstract}
We propose a novel framework for modeling the interaction between graphical structures and the natural language text associated with their nodes and edges. 
Existing approaches typically fall into two categories. On group ignores the relational structure by converting them into linear sequences and then utilize the highly successful Seq2Seq models. The other side ignores the sequential nature of the text by representing them as fixed-dimensional vectors and apply graph neural networks. Both simplifications lead to information loss.

Our proposed method utilizes both the graphical structure as well as the sequential nature of the texts. The input to our model is a set of text segments associated with the nodes and edges of the graph, which are then processed with a transformer encoder-decoder model, equipped with a self-attention mechanism that is aware of the graphical relations between the nodes containing the segments. This also allows us to use BERT-like models that are already trained on large amounts of text.

While the proposed model has wide applications, we demonstrate its capabilities on data-to-text generation tasks. Our approach compares favorably against state-of-the-art methods in four tasks without tailoring the model architecture. We also provide an early demonstration in a novel practical application -- generating clinical notes from the medical entities mentioned during clinical visits.
\end{abstract}

\section{Introduction}
Many applications in natural language processing require modeling snippets of texts whose relationships are best represented by graphical structures. In other words, they can be viewed as graphical networks whose nodes and edges are labeled with textual descriptions, such as knowledge graphs~\citep{ji2020survey} or the dialogue states in goal oriented dialog systems~\citep{chen2017survey,gao2018neural}.

One exemplar task is the data-to-text (D2T) generation, which aims at generating textual descriptions of input structured data. This task has attracted increasing research interest, accordingly several public datasets and challenges with different levels of complexity have been created. {\it WikiBio} \citep{wikibio} and {\it E2E} \citep{e2e} have relatively simple input structure consisting of key-value pairs, where the goal is to create a snippet that describes associated biographies and restaurants respectively. {\it AGENDA} \citep{GraphWriter} and {\it WebNLG} \citep{webnlg}, on the other hand, have more complex input graphs where the goal is to generate associated scientific abstracts, and natural language descriptions of given relations, respectively.

Current techniques for tasks such as D2T generally fall into two categories. On one side, the input graph is converted into a flat sequence, on which sequence-to-sequence (Seq2Seq) models~\citep{sutskever2014sequence} are employed. For example on the WebNLG dataset, where the graph relations are represented as (head, predicate, tail) tuples, one approach flattens the graph by concatenating the tuples, where the heads, predicates and tails are separated by special tokens~\citep{harkous2020have,webnlg,ferreira2019neural}. In a more elaborate approach, the input tree is traversed in a preferred order, and the branching structure is encoded using brackets~\citep{moryossef2019step}. While flattening the graph has been shown to be effective for modeling simple structures, the resulting sequences are not unique. The order of the tuples are often arbitrary and nodes that belong to multiple tuples may be represented multiple times in the sequence. In short, these methods do not utilize the relational structure effectively.

An alternative direction~\citep{GraphWriter} focuses on the graphical structure and exploits recent advances in graph neural network (GNN) architectures~\citep{wu2020comprehensive}. The textual descriptions are represented as fixed-dimensional vectors, for example by using the last state of a recurrent neural network (RNN). The node embeddings are initialized using these fixed vectors and the structure is modeled using a transformer-based GNN~\citep{velickovic2018graph}. Since the texts are only utilized for initializing the node embeddings, the textual information is not fully retained and they do not generalize well. This weakness could be mitigated by resorting to copy mechanism where the decoder is allowed to copy portions of the input~\citep{pointernetwork}.

In this work we propose a novel approach that utilizes the graphical structure of the data, while having access to the full sequences of text associated with graph components. We utilize the transformer-based models~\citep{Transformer}, whose self-attention layers and multi-head attention mechanism have proven their versatility in modeling both the natural language~\citep{devlin2018bert, brown2020language} and graphical structures~\citep{velickovic2018graph}.
After a preprocessing step that moves the original predicate texts (edge annotations) to newly added nodes, the new graph, whose predicates are now from a small closed set, is processed by our model.
Our model,  which can be viewed as an extension to the original transformer, is capable of processing multiple text segments (each corresponding to one node), while also utilizing the relations between the segments using a segment-wise attention mechanism. In another perspective, our model can be viewed as an attention-based graph neural network, where higher order relations between nodes (represented as text segments in the input) are utilized by stacking multiple attention layers in the transformer model.

Our model has several advantages to previous methods for data-to-text generation tasks.
First, our method does not require linearizing the input structure and can, hence, better incorporate the relational information of the data, whose importance increases with the complexity of the input structure and the scarcity of the training data.
Second, compared to previous GNN-based methods like Graph-Writer~\citep{GraphWriter}, we use a single transformer model for encoding both the relational information and the textual labels, eliminating the need for auxiliary encoders to encode textual labels to fixed-dimensional vectors (losing information), or relying on copy mechanism to generalize to unseen terms.
Third, compared to the more traditional pipeline methods~\cite{shen2019pragmatically,moryossef2019step,e2e}, our model is fully end-to-end and does not require additional steps such as content selection, delexicalization, etc.
Fourth, our architecture choices allow us to initialize the proposed model with a BERT-like~\citep{devlin2018bert} pretrained masked language model, such as XLNet~\citep{XLNET} in our experiments, to take advantage of unsupervised training on large corpora and the resulting improvements in generalization.

\section{Methods}
\subsection{Overview and task definition}
\begin{figure*}[t!]
\centering
    \begin{subfigure}[c]{0.25\textwidth}
        \includegraphics[width=\textwidth]{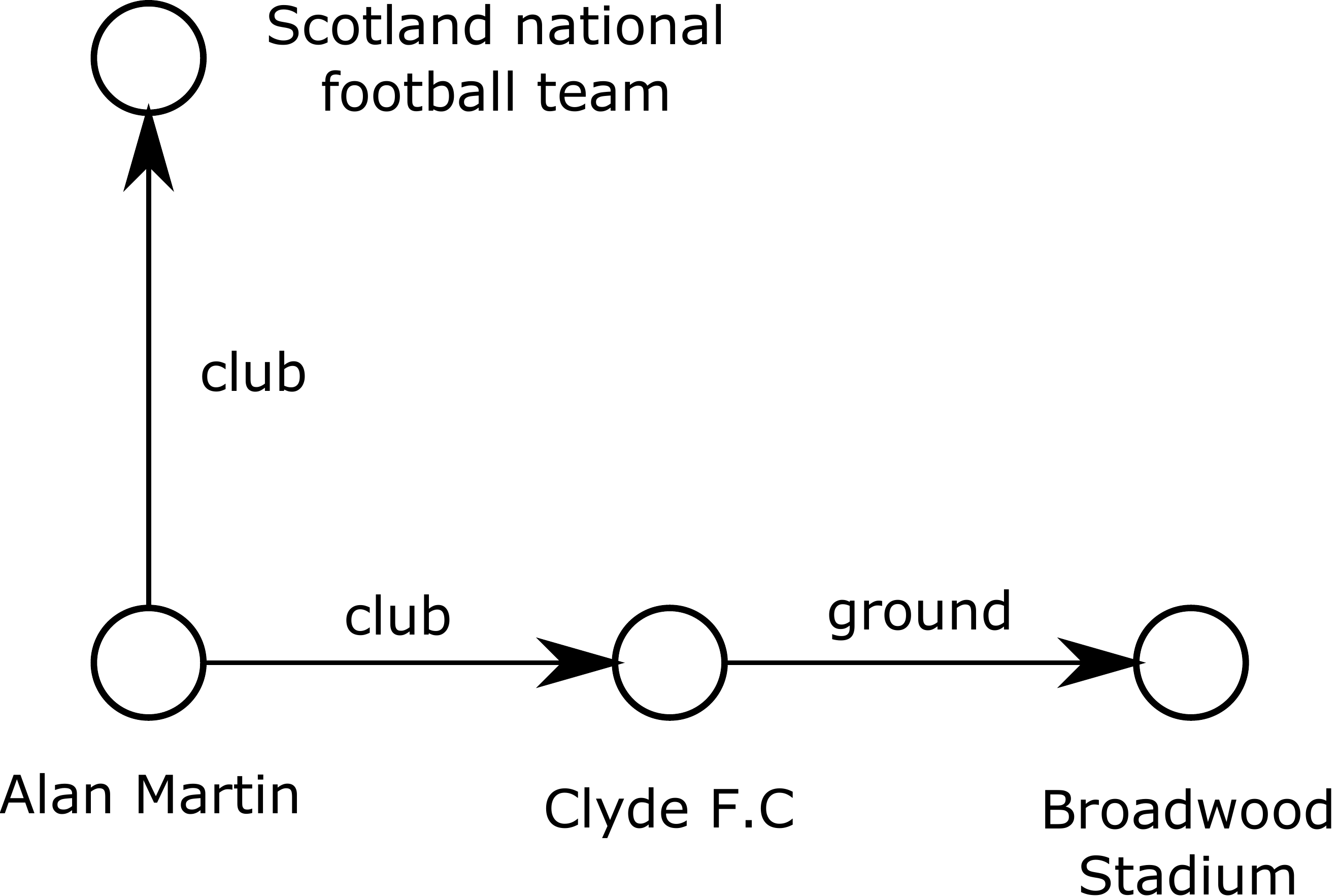}
        \caption{}
        \label{fig:graph_original_format}
    \end{subfigure}%
    \hspace{5mm}
    \begin{subfigure}[c]{0.53\textwidth}
        \includegraphics[width=\textwidth]{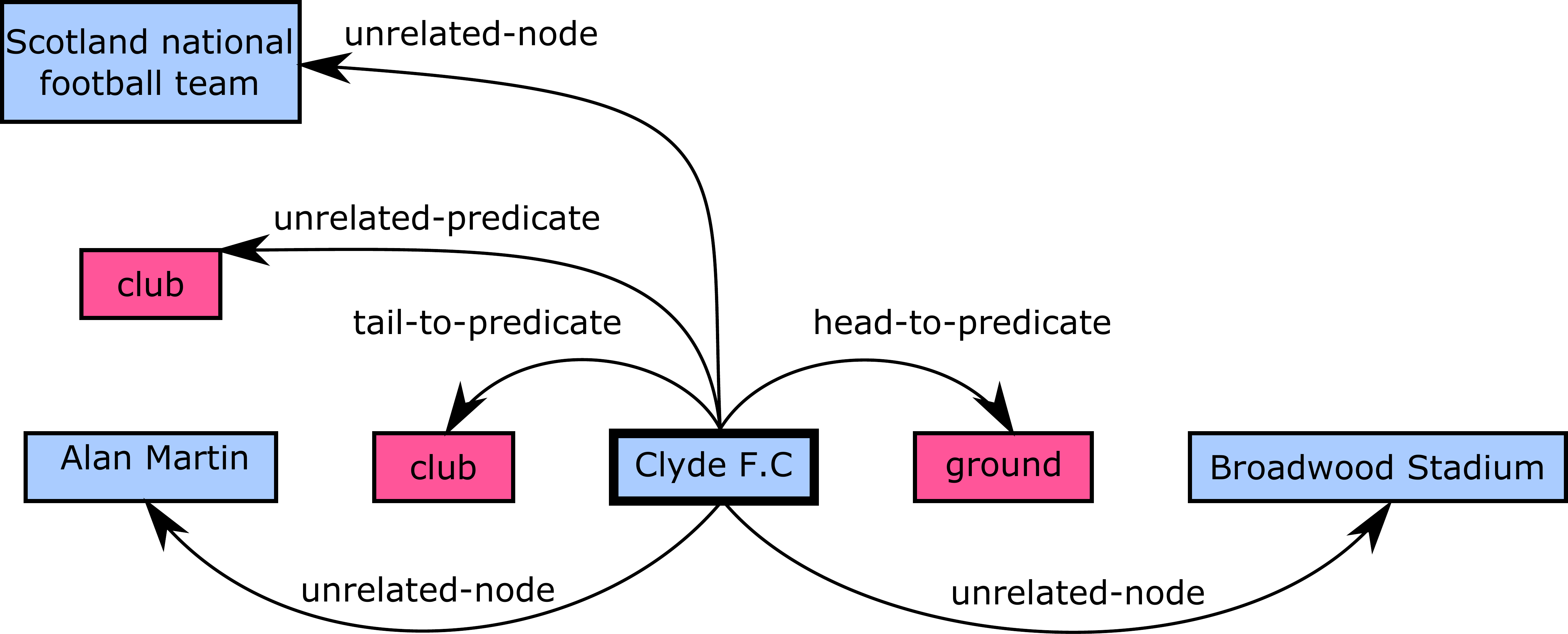}
        \caption{}
        \label{fig:graph_transformed_format}
    \end{subfigure}
    \caption{The plot on the left (a) depicts a graph example in its original format, where the predicates can have text labels from an open set. The plot on the right (b) shows the transformed version, where we have only depicted the relations originating from one specific node: ``Clyde F.C". The new graph is a network of text segments, each segment either being of the type ``node" or ``predicate", colored in blue and pink respectively.}
    \label{fig:graph_transform}
\end{figure*}

A standard graph structure can be described as a collection of tuples of the form -- (head, predicate, tail) -- where the heads and the tails collectively form the nodes in the graph and predicates correspond to the labels of the edges. 
The scope of our work involves graphs whose nodes, and possibly edges, are annotated with text labels.
Broadly, our goal is to infer the text associated with one or more target nodes, given the rest of the information in the graph. For example, in D2T tasks the goal is to infer the text that describes a given input graph. In this case, we regard the target text as the label for an extra dummy node. In general, the target nodes could be any subset of nodes in the graph.

The input graphs for different tasks may vary widely in format and characteristics. Often, we need to transform them into a structure that is compatible with our framework (or is more efficient).
If the predicates are text sequences from an open set (or a large closed set), then we convert the input graph into its equivalent Levi graph~\citep{levi1942finite}, similar to the preprocessing step also used by \citet{beck2018graph}. This is done by adding a new node for each tuple, having the predicate as the text associated to the new node, effectively breaking the original tuples such as (head, predicate, tail) into two new tuples -- (head, "\textit{head-to-predicate}", predicate) and (predicate, ``\textit{predicate-to-tail}", tail). We refer to Figure~\ref{fig:graph_transform} for an illustration of such a transformation.
After our transformation, every node, including the newly added ones, corresponds to a text snippet, which we call a \textit{text segment}, and the edges between them (the new predicates) form a closed categorical set. For the rest of the paper, unless specified, we are always referring to the transformed version of the graph.

More formally, given a set of $N_X$ text segments $X=\{\sS_i: 1 \le i \le N_X\}$ as source, our goal is to predict the set of $N_Y$ target text segments $Y=\{\sS_i: N_X < i \le N_X+N_Y\}$.
Each text segment is itself a sequence of tokens, where the $t$-th token for the $i$-th segment is notated as $\sS_{i,t}$.

Each segment also has a categorical segment type (out of $K$ possible types), notated as $B=(b_1, ..., b_{N_X+N_Y})$. For example, one can allocate two segment types ``\textit{node}'' and ``\textit{predicate}'', differentiating segments that correspond to the original nodes from the ones added to replace the original predicates in the Levi graph transformation.

We are also given a closed set of predicates $Q$ (either original predicates or new ones the after graph transformation), and a set of tuples:
\begin{equation*}
\begin{split}
T = \{&(\sS_{\mathrm{head}_i}, q_i, \sS_{\mathrm{tail}_i}): 1 \le i \le N_T;\\
&\sS_{\mathrm{head}_i}, \sS_{\mathrm{tail}_i} \in X \cup Y;  q_i \in Q\}.\\
\end{split}
\end{equation*}

We define an additional set of relations $L$, that supplements $Q$ to cover segment pairs that do not form any tuple. $L$ can have members such as ``\textit{same-segment}" for relations between each segment and itself, or ``\textit{unrelated}" for two unrelated ones. 
The underlying relational structure between the segments can then be represented by an adjacency matrix $G\in (Q\cup L) ^{(N_X+N_Y)\times(N_X+N_Y)}$, where:
\begin{equation*}
G_{i,j}=
\begin{cases}
  q, & \text{if}\ \exists q :(\sS_i, q, \sS_j) \in T \\
  l_{i,j} \in L, & \text{otherwise}.
\end{cases}
\end{equation*}

We should note that we have assumed each pair of segments can have at most one predicate in $T$, and the relations can be directional (i.e. it is possible that $G_{i,j} \neq G_{j,i}$).

Our goal is to estimate $P(Y \mid X, B, G)$, where we call the set of variables ($X$, $Y$, $G$, $B$) the data schema. Characteristics of our model and its architecture are provided Section \ref{sec:model}. We should also note that the choice of the supplementary relations $L$ and the segment types ($b_i$s) are design decisions and can vary across tasks. We have provided details of such design choices for our experiments Appendix \ref{sec:appendix-datasets}.

While D2T tasks only require inferring one target segment, our approach is more general and can infer multiple target segments, which may have other applications such as completing the missing nodes in knowledge graphs. Although it should be noted that due to the autoregressive nature of the generation process, each target segment can observe its relations only to the previously generated segments (in addition to the source segments).

\subsection{Model architecture}
\label{sec:model}
\begin{figure*}[t!]
    \centering
    \begin{subfigure}[c]{0.66\textwidth}
        \includegraphics[width=\textwidth]{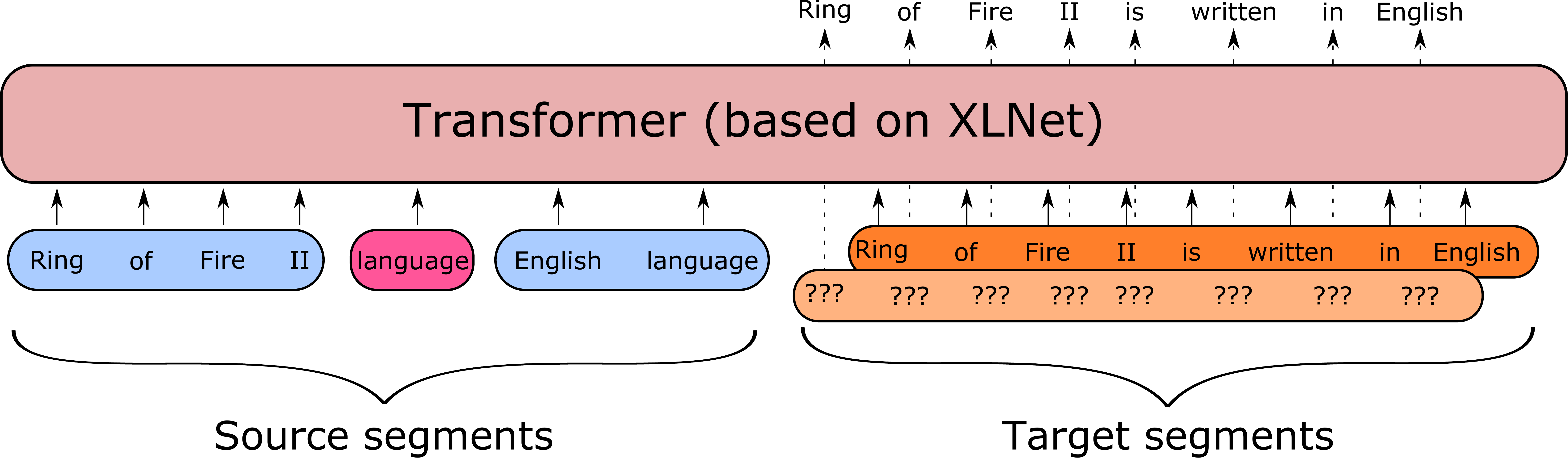}
        \caption{}
    \end{subfigure}%
    \hspace{5mm}
    \begin{subfigure}[c]{0.25\textwidth}
        \includegraphics[width=\textwidth]{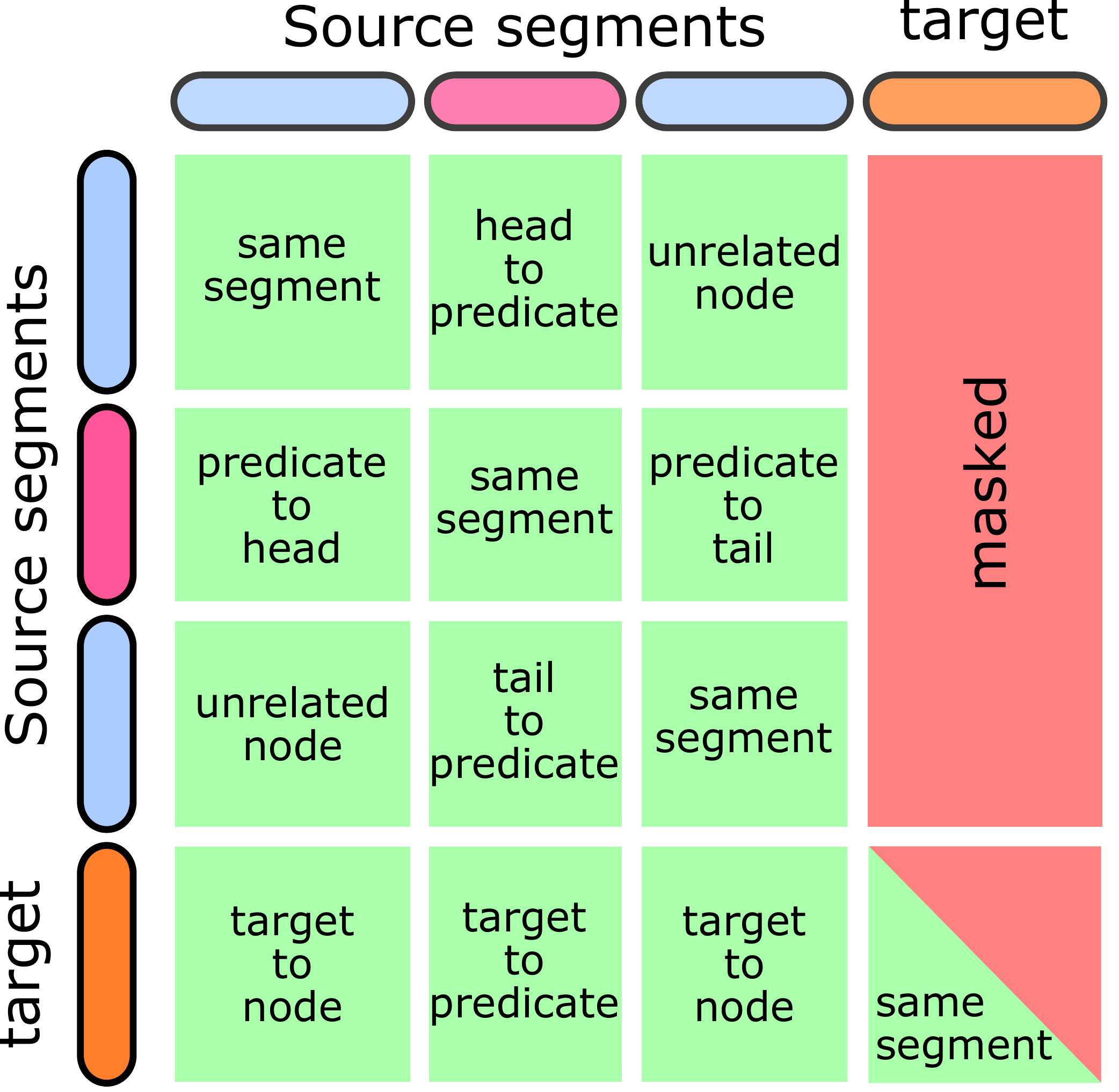}
        \caption{}
    \end{subfigure}
    \caption{(a) R2D2 Architecture based on the XLNet model. The input is the concatenation of all text segments, where the blue, pink and orange segments correspond to node, predicate and target segment types. The model predicts the target tokens autoregressively, using a two-stream self-attention mechanism, with a \textit{query-stream} colored in lighter orange passing a query "???" token, while the darker colored \textit{content-stream}, passes the target tokens (ground truth tokens during the training or the previously decoded tokens during the inference) as context for generating future tokens.
    (b) The self-attention between segment pairs. Each row corresponds to one segment as query, attending to other segments (columns) as keys. The colors show if the key is visible (green) or hidden (red) to the query, while the cell labels indicate the relation type, which influences the segment-wise attention scores.}
    \label{fig:model_arch}
\end{figure*}

We model $P(Y \mid X, B, G)$ with a conditional autoregressive model, parameterized by a transformer based on the XLNet-large architecture~\citep{XLNET}.
XLNet is a masked language model used for unsupervised pretraining, a variant in the family of BERT~\citep{devlin2018bert} style models, with the primary distinction of having an autoregressive decoding mechanism (opposed to the parallel decoding in other BERT variants).
To enable this, for each target token two parallel streams are passed, one is given a special query symbol as input and predicts the real token, while the other is given the real token and processes it as context for decoding future tokens.
See Figure \ref{fig:model_arch} for an overview.

The autoregressive nature of XLNet is desirable for our framework, as we are also interested in using our model to decode target segments autoregressively. On the other hand, compared to entirely autoregressive models such as GPT-2~\citep{radford2019language}, XLNet has the advantage of applying a bidirectional attention over the source segments.

Another important advantage of XLNet is the use of relative segment embedding, in contrast to the absolute segment embedding employed in other BERT variants. To avoid confusion, we should remind that segment embedding and positional embedding are two different concepts; the former helps attending to the entirety of a segment (node), whereas the latter attends at the token level. The relative segment embedding is a more suitable approach for modeling texts in complex graphical structures, as explained further in Section~\ref{sec:attention}.

The input to our model consists of the concatenation of all source and target segments.
Any concatenation order would be fine, as the model is invariant to the order of the segments. However, in case of having multiple target segments, an order should be given for the autoregressive generation process. 
The model does not ignore the boundary of the segments nor the inter-segment relations, as these aspects are modeled through the relative attention mechanisms implemented in XLNet, which is further explained in Section~\ref{sec:attention}.

While we use XLNet's embedding table to embed the sequence tokens, we also learn another embedding table for the segment types. The segment type ($b_i$) and the token ($\sS_{i,t}$) embeddings are then summed to provide the final representation of each input token fed to the transformer.

\subsection{Multi Segment Self-Attention}
\label{sec:attention}
Pretrained transformer models such as BERT usually allow having multi-segment inputs (usually limited to two segments), to help with tasks such as question answering where the context passage and the question can be marked as separate segments.

BERT and most of its other variations employ a global segment embedding for each of the two segment types (e.g. context and question) and add it to the input embedding to let the model know which tokens belong to which segment.
In contrast, XLNet uses a relative segment embedding approach, where the attention score between two positions is affected by whether they both belong to the same segment or not, enabling each transformer cell to decide which segment to attend to.

\begin{figure}[h]
    \centering
    \includegraphics[width=0.90\linewidth]{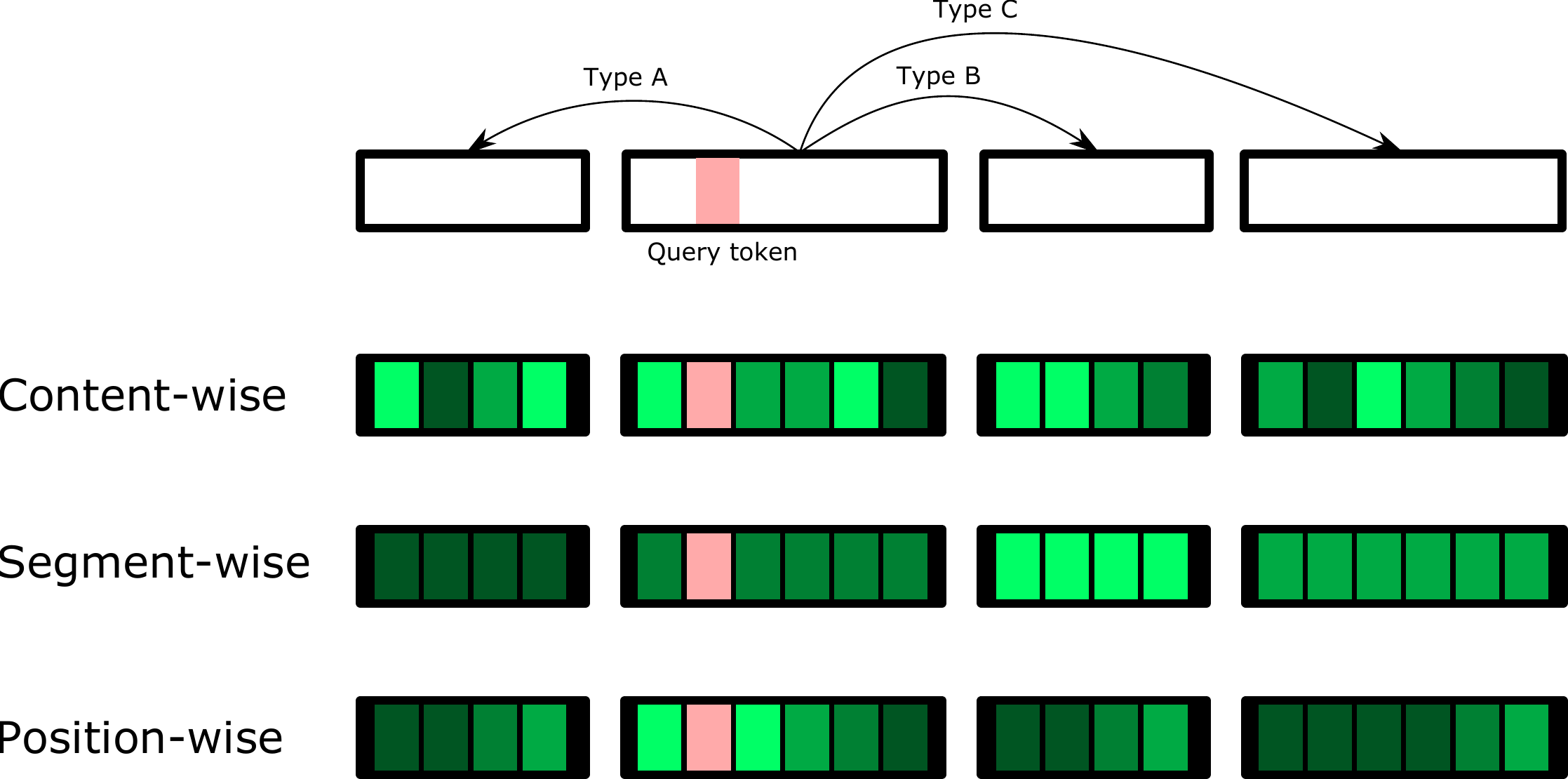}
    \caption{Multi-segment attention. The first row shows the structure of the text segments. The next three rows illustrate attention scores for each of the content-wise, segment-wise and position-wise components respectively. The shade of the green reflects the magnitude of the score.
    Note that every token of each segment shares the same segment-wise score, while for the position-wise attention the three external segments share the same pattern of attention from right to left.}
    \label{fig:attentiontypes}
\end{figure}

In this work, we extended the relative segment embedding to more than two segments, where the inter-segment attention depends on the type of relation between pairs of segments.
The attention score between any two positions has three components: 1) content-wise, 2) position-wise and 3) segment-wise. The scores of all three are summed to compute the final attention score. Figure~\ref{fig:attentiontypes} shows an example for the three attention components.

Consider we are calculating the attention score for the query token $i$ at position $t_i$ of the segment $\sS_i$ and the key token $j$ at position $t_j$ of segment $\sS_j$. Further assume at the current layer of the transformer, the query and key vector for $i$ and $j$ are $h_i^q$ and $h_j^k$ respectively.
The content-wise attention score is then calculated as:
\begin{equation}
a_{i,j}^\textrm{content} = \left( h^q_{i} + \phi_c \right)^T h^k_{j}.
\end{equation}

where $\phi_c$ is a learnable bias parameter.

To calculate the segment-wise attention, we learn a table of relation-type embedding $H \in \mathbb{R}^{(|Q|+|L|) \times d}$, where each row corresponds to one relation category and $d$ is the embedding dimension.
Assuming $\sigma_i$ and $\sigma_j$ are the segment indices for $\sS_i$ and $\sS_j$, the segment-wise attention score between $i$ and $j$ is derived as:
\begin{equation}
a_{i,j}^\textrm{segment} = \left( h^q_{i} + \phi_b \right)^T H_{G_{\sigma_i, \sigma_j}}.
\end{equation}

The positional attention score is computed using a table of relative positional embedding $R$, where $R_{\tau(i, j)}$ is the embedding for the relative position between $t_i$ and $t_j$ and $\tau(i, j)$ is the relative position of $t_i$ and $t_j$.
If the two positions are located in the same segment, then deriving $\tau(i,j)$ would be straightforward as the relative position is simply the positional difference. However if the two belong to different segments, then given the absence of an underlying global positioning, the relative position between the two requires a new definition.
We calculate the relative position between the tokens $i$ and $j$ as if their respective segments are concatenated as $[\sS_j, \sS_i]$. 
More formally:
\begin{equation}
\tau(i, j)=
\begin{cases}
  t_i - t_j, & \text{if}\ \sigma_i = \sigma_j\\
  t_i-t_j+|\sS_j|, & \text{otherwise}.
\end{cases}
\end{equation}

The positional attention score can be then derived as:
\begin{equation}
a_{i,j}^\textrm{pos} = \left( h^q_{i} + \phi_p \right)^T R_{\tau(i,j)}.
\end{equation}

We should note that the reason we are using relative positional embedding in the first place is simply for compatibility with the pretrained XLNet model, otherwise our model could also utilize absolute positional embedding, where the leftmost token of each segment is viewed as having the absolute position 1, and tokens from different segments sharing the same absolute position are differentiated based on the content-wise and segment-wise attentions. 

The final attention score is simply the sum of all three components:
\begin{equation}
a_{i,j} = a_{i,j}^\textrm{content} + a_{i,j}^\textrm{segment} + a_{i,j}^\textrm{pos}.
\end{equation}
\section{Experiments}
\subsection{Tasks}
We evaluated our model on a range of tasks related to data-to-text generation, which includes WikiBio~\citep{lebret2016neural}, with $728$k infobox-biography pairs from English Wikipedia, E2E challenge~\citep{e2e} with $50$k examples in the restaurant reservation domain, AGENDA~\citep{GraphWriter}, with $40$k scientific abstracts, each supplemented with a title and a matching knowledge graph, and WebNLG~\citep{webnlg}, containing $25$k graph-text pairs, with graphs consisting of tuples from DBPedia and text being a verbalization of the graph. On all datasets we repeated our experiments (including the training process) three times and summarized our results by reporting the mean and variance of each metric.

The AGENDA and WebNLG tasks involve more complex graphs as their source, presented as general sets of tuples of the form (head, predicate, tail). The two, however, differ on the set of their permitted predicates: AGENDA comes with a closed set of seven predicate types, while the WebNLG predicates are arbitrary texts, comprising an open set. For both tasks, we grant one text segment per node, whereas for the predicates, we simply use the original seven possible predicates in AGENDA as distinct categories, while for WebNLG, we perform the Levi graph transformation and allocate additional text segments for the predicates, connecting them to their associated head and tail segments as illustrated in Figure~\ref{fig:graph_transform}. See Appendix \ref{sec:appendix-datasets} for task-specific data/graph preparation details. 

\subsection{Baselines}
We compared our method with the current state-of-the-art models on each task (chosen based on the BLEU-4 score) and performed ablation studies to understand the impact of different components of our model.

We did ablation studies with four variants of our model. In the "scratch" variant, the model architecture was kept the same as XLNet but the parameters were initialized with random values and learned from scratch. In the "flattened" variant, the input structure was converted into a sequence by separating the different components of the relations -- heads, tails, predicates, keys, values -- with special tokens. In the "SF" variant, we both trained from scratch and flattened the input. Finally, the regular R2D2 is our proposed variant, utilizing both the pretrained weights and the graphical structure.

Current state-of-the-art model on WikiBio is the Structure Aware model introduced by \citet{liu2018table}. Their method flattens the input into two parallel sequences of keys and values. They are then processed in a Seq2Seq framework, where the LSTM decoder is augmented with a dual attention mechanism. They compute and sum two attention scores for each of the two sequences.
Current state-of-the-art model on E2E is the Pragmatically Informative model of \citet{shen2019pragmatically}, which is a Seq2Seq model whose input is delexicalized. During inference, they combine multiple scores in a beam search.
The state-of-the-art model on AGENDA, the GraphWriter~\citep{GraphWriter}, uses an RNN encoder to map each text label into a fixed-dimension vector, which are then fed to a GNN model to process the graph. A decoder uses the output of the GNN as context and generates the target sequence. The decoder can also access the input text via a copy mechanism.
Current state-of-the-art for WebNLG is the recent DataTurner model proposed by \citet{harkous2020have}. They fine-tune the pre-trained GPT-2 model~\citep{radford2019language}, which is shared between the encoder and the decoder. The input to the model is a flattened version of the graph.

\subsection{Results}
\begin{table}[h]
\hspace{-0.3cm}
\centering
\begin{tabular}{||c | c c||} 
 \hline
 Method &  BLEU-4 & ROUGE-4\\ 
 \hline\hline
  StructureAware* & 44.89 \footnotesize{$\pm 0.33$} & 41.65 \footnotesize{$\pm 0.25$}\\
 \hline
  R2D2 & {\bf 46.23} \footnotesize{$\pm 0.15$} & 45.10 \footnotesize{$\pm 0.28$} \\
  R2D2 (Flattened) & 45.62 \footnotesize{$\pm 0.73$} & {\bf 45.56} \footnotesize{$\pm 0.33$} \\ 
  R2D2 (Scratch) &  44.33 \footnotesize{$\pm 0.53$} & 44.22 \footnotesize{$\pm 0.33$} \\ 
  R2D2 (SF) &  44.01 \footnotesize{$\pm 0.65$} & 43.89 \footnotesize{$\pm 0.38$} \\ 
 \hline 
 \multicolumn{3}{||c||}{Reduced  training set (R2D2)} \\
 \multicolumn{3}{||c||}{(number of samples / percent of training data)} \\ \hline 
  R2D2 (70k / 12\%) & 44.40 \footnotesize{$\pm 0.12$} & 44.25 \footnotesize{$\pm 0.07$} \\ 
  R2D2 (5k / 1\%) & 37.19 \footnotesize{$\pm 0.15$} & 39.08 \footnotesize{$\pm 0.43$} \\
  R2D2 (.5k / 0.1\%) & 33.00 \footnotesize{$\pm 0.82$} & 34.10 \footnotesize{$\pm 0.42$} \\  
 \hline
\end{tabular}
\caption{\label{tab:wikibio}Results on the WikiBio task. State-of-the-art~\citep{liu2018table} is notated by *.}
\end{table}

The results of our experiments on WikiBio are reported in Table~\ref{tab:wikibio}, which show that our model outperforms the state-of-the-art~\citep{liu2018table} with a considerable margin on both BLEU-4 and ROUGE-4 scores.
The results for the E2E experiments are reported in Table~\ref{tab:e2e}, which show that our model is comparable to the state-of-the-art~\cite{shen2019pragmatically} model; we observe a small improvement in terms of NIST, METEOR, ROUGE-L and CIDEr, but with a slight reduction in BLEU-4 score.
On WebNLG task, as reported in Table~\ref{tab:webnlg}, our model performs comparably to the state-of-the-art model~\citep{harkous2020have}, achieving a small improvement in BLEU-4 score and a slight reduction on the METEOR metric.
Our results on the AGENDA task, reported in Table~\ref{tab:agenda}, show that our model outperforms the baseline (GraphWriter)~\citep{GraphWriter} model by a considerable margin. 
Full results of our few-shot experiments, which includes all variants of R2D2, is available in Appendix \ref{sec:appendix-fewshot}.
\begin{table*}[h]
\centering
\renewcommand\tabcolsep{4pt}
 \begin{tabular}{||c | c c c c c||} 
 \hline
 Method &  BLEU-4 & NIST & METEOR & ROUGE-L & CIDEr\\ 
 \hline\hline
  PragInfo* & \bf 68.60 & \bf 8.73 & 45.25 & \bf 70.82 & 2.37 \\
  \hline  
   R2D2 & 67.70 \footnotesize{$\pm 0.64$}  & 8.68 \footnotesize{$\pm 0.10$} & \bf 45.85 \footnotesize{$\pm 0.18$} & 70.44 \footnotesize{$\pm 0.32$} & \bf 2.38 \footnotesize{$\pm 0.04$}\\
 R2D2 (Flattened) & 65.73 \footnotesize{$\pm 0.97$}  & 8.48 \footnotesize{$\pm 0.13$} & 45.29 \footnotesize{$\pm 0.24$} & 70.21 \footnotesize{$\pm 0.73$} &  2.32 \footnotesize{$\pm 0.05$}\\
  R2D2 (Scratch) & 52.25 \footnotesize{$\pm 2.45$}  & 7.31 \footnotesize{$\pm 0.16$} & 37.40 \footnotesize{$\pm 1.05$} & 60.13 \footnotesize{$\pm 1.76$} &  1.34 \footnotesize{$\pm 0.08$}\\
  R2D2 (SF) & 52.25 \footnotesize{$\pm 3.56$}  & 7.24 \footnotesize{$\pm 0.21$} & 36.66 \footnotesize{$\pm 2.10$} & 58.97 \footnotesize{$\pm 1.91$} &  1.31 \footnotesize{$\pm 0.07$}\\
 \hline
 \multicolumn{6}{||c||}{Reduced training set (R2D2) (number of samples / percent of training data)} \\ \hline 
 R2D2 (4k / 10\%) &  67.01 \footnotesize{$\pm 0.65$}  & 8.65 \footnotesize{$\pm 0.06$} & 45.26 \footnotesize{$\pm 0.30$} & 69.48 \footnotesize{$\pm 0.60$} &  2.29 \footnotesize{$\pm 0.06$}\\
 R2D2 (1k / 2\%) &  63.93 \footnotesize{$\pm 1.13$}  & 8.46 \footnotesize{$\pm 0.11$} & 43.90 \footnotesize{$\pm 1.18$} & 66.95 \footnotesize{$\pm 0.76$} &  2.15 \footnotesize{$\pm 0.07$}\\
 R2D2 (.5k / 1\%) & 64.26 \footnotesize{$\pm 1.22$}  & 8.48 \footnotesize{$\pm 0.10$} & 43.48 \footnotesize{$\pm 0.85$} & 66.74 \footnotesize{$\pm 1.14$} &  2.12 \footnotesize{$\pm 0.05$}\\
 \hline
\end{tabular}
\caption{\label{tab:e2e}Results on the E2E task. State-of-the-art~\citep{shen2019pragmatically} is notated by *.}
\end{table*}

\begin{table}[h]
\centering
 \begin{tabular}{||c | c c||} 
 \hline
 Method &  BLEU-4 & METEOR\\ 
 \hline\hline
  DataTurner* & 52.9 & {\bf 41.9}\\
 \hline
  R2D2 & {\bf 53.77} \footnotesize{$\pm 0.86$} & 41.30 \footnotesize{$\pm 0.36$}\\ 
  R2D2 (Flattened) & 53.26 \footnotesize{$\pm 1.41$} & 40.04 \footnotesize{$\pm 0.47$}\\
  R2D2 (Scratch) &  33.25 \footnotesize{$\pm 2.16$} & 25.42 \footnotesize{$\pm 0.17$}\\  
  R2D2 (SF) &  32.62 \footnotesize{$\pm 0.07$} & 24.40 \footnotesize{$\pm 0.01$}\\  
 \hline
 \multicolumn{3}{||c||}{Reduced training set (R2D2)} \\ 
 \multicolumn{3}{||c||}{(number of samples / percent of training data)} \\ \hline 
  R2D2 (1.8k / 10\%) &  52.98 \footnotesize{$\pm 0.40$} & 40.80 \footnotesize{$\pm 0.42$}\\  
   R2D2 (360 / 2\%) &  47.58 \footnotesize{$\pm 0.41$} & 38.12 \footnotesize{$\pm 0.20$}\\  
  R2D2 (180 / 1\%) &  42.86 \footnotesize{$\pm 0.74$} & 36.23 \footnotesize{$\pm 1.09$}\\  
 \hline
\end{tabular}
\caption{\label{tab:webnlg}Results on the WebNLG task. State-of-the-art~\citep{harkous2020have} is notated by *.}
\end{table}

\begin{table}[h]
\centering
 \begin{tabular}{||c | c c ||} 
 \hline
 Method &  BLEU-4 & METEOR\\ 
 \hline\hline
  {GraphWriter*} & 14.3 \footnotesize{$\pm 1.01$}& 18.8 \footnotesize{$\pm 0.28$}\\
 \hline
  R2D2 & {\bf 17.30} \footnotesize{$\pm 0.20$} & {\bf 21.82} \footnotesize{$\pm 0.15$} \\
  R2D2 (Flattened) &  16.17 \footnotesize{$\pm 0.72$} & 20.36 \footnotesize{$\pm 0.38$} \\
  R2D2 (Scratch) &  10.60 \footnotesize{$\pm 0.22$} & 15.71 \footnotesize{$\pm 0.09$} \\
  R2D2 (SF) &  9.87 \footnotesize{$\pm 0.75$} & 14.63 \footnotesize{$\pm 0.18$} \\
  \hline 
    \multicolumn{3}{||c||}{Reduced training set (R2D2)} \\ 
 \multicolumn{3}{||c||}{(number of samples / percent of training data)} \\ \hline 
  R2D2 (5k / 13\%) & 16.83 \footnotesize{$\pm 0.18$} & 21.71 \footnotesize{$\pm 0.24$} \\
 R2D2 (1k / 2.6\%) &  14.60 \footnotesize{$\pm 0.59$} & 19.03 \footnotesize{$\pm 0.80$} \\
  R2D2 (.5k / 1\%) &  13.00 \footnotesize{$\pm 0.38$} & 17.59 \footnotesize{$\pm 0.19$} \\
 R2D2 (.1k / 0.26\%) & 10.61 \footnotesize{$\pm 1.77$} & 15.56 \footnotesize{$\pm 1.20$} \\
  \hline
\end{tabular}
\caption{\label{tab:agenda}Results on the AGENDA task. State-of-the-art~\citep{GraphWriter} is notated by *.}
\end{table}

\section{Discussion}
The main benefits of our model come from three aspects: 1) encoding the graphical structure explicitly using the segment-wise attention mechanism; 2) concurrently modeling both the graphical structure and the text labels within a single transformer architecture, where the encoded text representations retain their full sequential information and length; 3) utilizing a pretrained language model like XLNet, trained on large amounts of data.


\subsection{Pretraining and few shot learning}
Our ablation studies show that without using the pretrained model, experiments on relatively small datasets such as AGENDA, E2E and WebNLG performed significantly worse, while this performance gap is much smaller for larger datasets such as WikiBio. Note, for experimental efficiency, we  used the same large architecture for all tasks and smaller models for the smaller tasks may have avoided overfitting and improved performance.

The baseline model for the WebNLG task, the DataTurner model~\citep{harkous2020have}, also uses a pretrained  GPT-2~\citep{radford2019language} model. GPT-2 is a fully autoregressive model, as such its information flow, even on the source input, is unidirectional. In contrast our XLNet model has bidrectional information flow for the source input.

The results of our few-shot experiments show that the pretrained model can achieve relatively good results with reduced training data. More specifically, training on about $10\%-13\%$ of the data from Wikibio (70k samples), E2E (4k samples), AGENDA (5k samples) and WebNLG (1.8k samples), led to a BLEU-4 score reduction of only 3.9\%, 4\%, 2.7\%, and 0.3\%, respectively, validating that low-resource data-to-text tasks can utilize transfer learning from large corpora.

\subsection{Joint Encoding of Graph and Text}
To understand the impact of our model explicitly encoding the graphical structure, we conducted ablation experiments by flattening the structured input into a linear sequence. This can be viewed as a regular Seq2Seq task. The loss of structural information resulted in $.2\%$, $2.9\%$, $6.5\%$, $0.9\%$ BLEU-4 score decrease on WikiBio, E2E, AGENDA, and WebNLG tasks respectively. As expected, AGENDA and WebNLG, which have more complex graph structures, showed a larger drop in performance when flattened, compared to WikiBio. More importantly, our ablation study shows that utilizing the graph structure can boost the performance independent of whether pretrained weights were used and the two have a complementary effect on the performance gain.

One surprising observation was that the graphical structure benefited the relatively simple E2E task too, where perhaps the structural information helped offset the data scarcity (compared to WikiBio).
We further investigated this phenomenon by repeating the few shot experiments on the flattened version of the model. The performance gain from utilization of the graphical structure was indeed more significant in several few shot experiments, such as WebNLG where reducing the training data to 1800 samples led to 1.4\% and 5.7\% drops in BLEU score for the main variant and the flattened version respectively.

\subsection{End-to-end Learning}
State-of-the-art models for certain tasks have multiple steps such as delexicalization and second pass rescoring, as mentioned earlier~\cite{shen2019pragmatically,moryossef2019step,e2e}. Our model on the other hand is purely end-to-end without the hassles of delexicalization, copy mechanism, or other interventions. Consequently, all model components including those devoted to textual description and structural information can be trained simultaneously, affording an opportunity to achieve better overall performance.

\subsection{Computational complexity}
The computational cost of R2D2 would be similar to that of regular transformers, requiring $O(n^2)$ dot product computations per layer, where n is the total length of all text segments concatenated.
Consequently, the framework would allow graphs that have less than $n_\mathrm{max}/n_\mathrm{seg}$ number of nodes, where $n_\mathrm{max}$ is the maximum permitted sequence length for the transformer and $n_\mathrm{seg}$ is the average segment length. Accordingly, R2D2 is more suitable for tasks involving smaller graphs or local sub graphs.

\section{Applications in clinical note generation}
We applied our model to an early stage demonstration of a novel clinical application: generating clinical notes from extracted entities.
This is motivated by the need to improve the accuracy and efficiency of clinical documentation (e.g., \citep{liu2018learning}), which takes an inordinate amount of time from clinicians after patient visits.
With current capabilities of speech recognition and entity extraction, we could potentially extract the entities into a relevant data structure and then synthesize the clinical notes.
\begin{figure}[h]
    \centering
    \includegraphics[width=0.90\columnwidth]{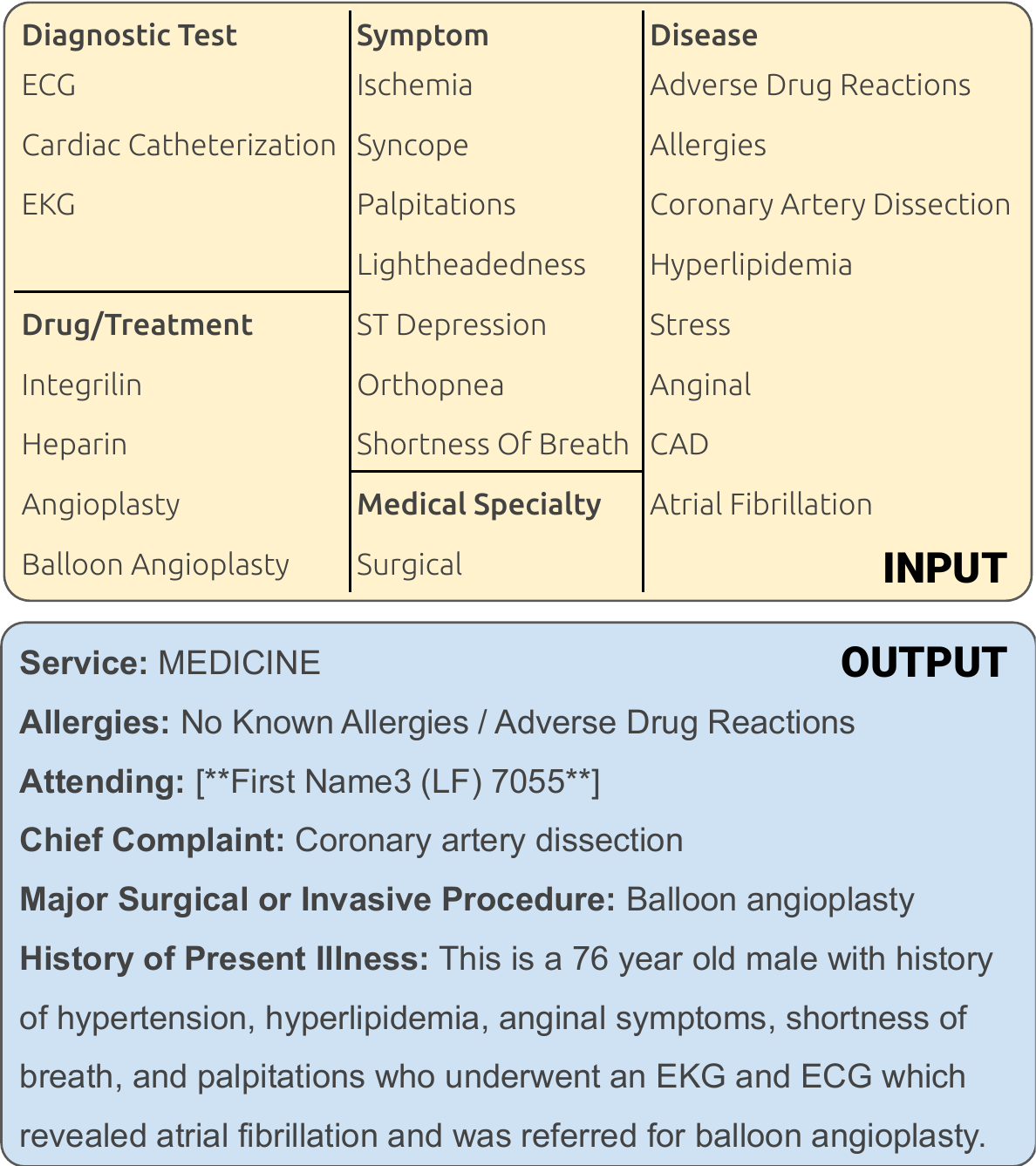}
    \caption{Clinical note generation with MIMIC-III data.}
    \label{fig:mimic}
\end{figure}

For this proof-of-concept demonstration, we used ICU discharge notes from the MIMIC-III dataset \citep{mimiciii}. Each section was chunked into text snippets of at most $250$ tokens. We then used an in-house entity extractor~\citep{Brown2013globally}
to extract medical key-value pairs, such as symptom, medication, condition, etc. (see Figure~\ref{fig:mimic}). We used these key-value pairs as the input graph and the text snippet as the target text.

Automatic evaluation results on our validation set ($5,765$ samples) show scores of BLEU-4 $18.43$, NIST $1.7486$, METEOR $0.1957$, ROUGE-L, $0.3340$, and CIDEr $0.0812$.
Figure~\ref{fig:mimic} shows the snippet of an inferred clinical note.
The model is able to capture information well from the input table. However when there is information in the target sequence for which the relevant entity/attributes are missing in the input, the model learns to hallucinate.
Similarly, when attributes are missing from the input table, such as symptom negation, severity, etc., the model is confused whether to generate "patient denied symptoms" or "patient reported symptoms".
Techniques such as forced copy mechanism might be necessary to safeguard against generating text that is not confidently supported by the input.
\section{Conclusion}
In this paper, we proposed a novel architecture for modeling the interaction between graphical structure where nodes (and possibly edges) contain descriptive text, and demonstrated performance gains over state-of-the-art models on the task of generating text from structured data.
The framework allows us to exploit pretrained XLNet encoder and achieve relatively good performance even with small amounts of training data.
Our ablation analysis showed that although using the pretrained model has a significant impact on the performance gain, utilizing the graphical information with the multi-level attention mechanism has a complementary effect and boosts the performance independent of whether pretrained weights were used. Our attempt to utilize this model in generating clinical notes hopefully inspires more work in this direction to solve a practical application of considerable importance.

\bibliography{r2d2}
\bibliographystyle{acl_natbib}

\appendix
\clearpage
\section{Dataset details and preparations}
\label{sec:appendix-datasets}
\subsection{WikiBio}
The WikiBio corpus \citep{lebret2016neural} comprises $728,321$ biography-infobox pairs from English Wikipedia (dumped in 2015). The task consists of generating the first sentence of a biography conditioned on the structured infobox. The average length of infoboxes and first sentences are $53.1$ and $26.1$ respectively. All tokens are lower-cased. 

We associate a text segment for each key and value, and a separate segment for the target sentence. Five relation types were defined: 1) from a key/value to its paired value/key; 2) from a key/value/target to every other key except if the two belong to a key-value pair; 3) from a key/value/target to every other value except if the two belong to a key-value pair; 4) from a key/value to the target (these are masked); 5) from any segment to itself.

\subsection{E2E}
The E2E challenge is based on a crowd-sourced dataset of 50k samples in the restaurant domain. The task is to generate a description of a restaurant from structured input in the form of key-value pairs. We used a graph structure identical to the one explained for the WikiBio dataset.

\subsection{AGENDA}
AGENDA dataset \citep{GraphWriter} comprises $40,720$ scientific abstracts, each supplemented with the title and a matching knowledge graph. The knowledge graphs consist of entities and relations extracted from their corresponding abstract. The dataset is split into $38,720$ training, $1,000$ validation, and $1,000$ test datapoints. The average lengths of titles and abstracts are $9.5$ and $141.2$ tokens respectively. The goal of the task is to generate an abstract given the title and the associated knowledge graph as the source. The input graphs in this dataset can form more complex structures compared to WikiBio and E2E, which are limited to key-value pairs.

The knowledge graphs consists of a set of ``\textit{concepts}" and a set of predicates connecting them. There are seven predicate categories, namely: ``Used-for", ``Feature-of", ``Conjuction", ``Part-of", ``Evaluate-for", ``Hyponym-for", and ``Compare".  Each concept is also assigned a \textit{class}, out of five possible classes: ``Task", ``Method", ``Metric", ``Material", or ``Other Scientific Term". We also incorporate the given title in the knowledge graph as a concept and give its own class of \textit{title-class}. Clearly, this new concept does not have any predicates connecting it to others.

We assign a text segment for each concept, class and target (the abstract), adding up to a total of three segment types. We also define a total of twelve types of relations, including the original seven predicate categories. The five supplementary relations are:
1) from a concept/class to its paired class/concept; 2) from a concept/class/target to every other concept unless the two are already linked; 3) from a concept/class/target to every other class except unless the two are already linked; 4) from a concept/class to the target (these are masked); 5) from any block to itself.

\subsection{WebNLG}
WebNLG dataset, curated from DB-Pedia, consists of sets of (head, predicate, tail) tuples forming a graphical structure and a descriptive target text verbalising them. The dataset consists of $25,298$ graph-text pairs, among which $9,674$ are distinct graphs (multiple targets for a graph produced by different crowdsourcing agents). In each graph there can be between one to seven tuples. The test data spans fifteen different domains, ten of which appear in the training data.

Both nodes (heads and tails) and the predicates are textual labels and the predicates are from an open set. As a result, we transform the graph to its Levi equivalent by allocating a new node per tuple with the predicate as its text label. 
This new graph has three types of text segments: nodes (original heads and tails), predicates (newly added nodes), and the target.
There are seven types of relations in the new graph: 1) from a node segment to all predicate segments, for which the node was a head; 2) from a node segment to all predicate segments, for which the node was a tail; 3) from each node/predicate to all predicates, except if the two were linked by a tuple; 4) from each node to all other nodes (including the ones that are linked by a tuple) and from each predicate to all nodes except if the two are linked by a tuple; 5) from target to all node segments; 6) from target to all predicate segments; 7) from a segment to itself.
\section{Complete Experiment results}
Full results of our experiments are provided in Table \ref{tab:fs-wikibio}, Table \ref{tab:fs-e2e}, Table \ref{tab:fs-webnlg} and Table \ref{tab:fs-agenda}, for WikiBio, E2E, WebNLG and AGENDA tasks respectively. For WebNLG and E2E we used the official evaluation scripts to compute the scores. We also used the popularly used ROUGE-1.5.5.pl to calculate the ROUGE-4 scores.

\label{sec:appendix-fewshot}
\begin{table*}[h]
\centering
\begin{tabular}{||c | c c||} 
 \hline
 Method &  BLEU-4 & ROUGE-4\\ 
 \hline\hline
  StructureAware* & 44.89 \footnotesize{$\pm 0.33$} & 41.65 \footnotesize{$\pm 0.25$}\\
 \hline
  R2D2 & {\bf 46.23} \footnotesize{$\pm 0.15$} & 45.10 \footnotesize{$\pm 0.28$} \\
  R2D2 (Flattened) & 45.62 \footnotesize{$\pm 0.73$} & {\bf 45.56} \footnotesize{$\pm 0.33$} \\ 
  R2D2 (Scratch) &  44.33 \footnotesize{$\pm 0.53$} & 44.22 \footnotesize{$\pm 0.33$} \\ 
  R2D2 (SF) &  44.01 \footnotesize{$\pm 0.65$} & 43.89 \footnotesize{$\pm 0.38$} \\ 
 \hline 
 \multicolumn{3}{||c||}{Reduced  training set (R2D2)} \\
 \multicolumn{3}{||c||}{(number of samples / percent of training data)} \\ \hline 
  R2D2 (70k / 12\%) & 44.40 \footnotesize{$\pm 0.12$} & 44.25 \footnotesize{$\pm 0.07$} \\ 
  R2D2 (5k / 1\%) & 37.19 \footnotesize{$\pm 0.15$} & 39.08 \footnotesize{$\pm 0.43$} \\
  R2D2 (.5k / 0.1\%) & 33.00 \footnotesize{$\pm 0.82$} & 34.10 \footnotesize{$\pm 0.42$} \\  
 \hline
\multicolumn{3}{||c||}{Reduced  training set (R2D2 Flattened)} \\
\multicolumn{3}{||c||}{(number of samples / percent of training data)} \\ \hline 
 R2D2 (Flattened) (70k / 12\%) & 44.44 \footnotesize{$\pm 0.18$} & 44.21 \footnotesize{$\pm 0.08$} \\ 
 R2D2 (Flattened) (5k / 1\%) & 36.39 \footnotesize{$\pm 0.21$} & 39.31 \footnotesize{$\pm 0.56$} \\
 R2D2 (Flattened) (.5k / 0.1\%) & 30.53 \footnotesize{$\pm 1.09$} & 31.60 \footnotesize{$\pm 0.98$} \\  
\hline 
\multicolumn{3}{||c||}{Reduced  training set (R2D2 - Scratch)} \\
\multicolumn{3}{||c||}{(number of samples / percent of training data)} \\ \hline 
R2D2 (Scratch) (70k / 12\%) & 43.16 \footnotesize{$\pm 0.34$} & 42.67 \footnotesize{$\pm 0.44$}\\ 
R2D2 (Scratch) (5k / 1\%) & 34.48 \footnotesize{$\pm 0.51$} & 33.40 \footnotesize{$\pm 0.79$}\\ 
R2D2 (Scratch) (.5k / 0.1\%) & 7.94 \footnotesize{$\pm 0.71$} & 28.00 \footnotesize{$\pm 0.82$}\\  
\hline 
\multicolumn{3}{||c||}{Reduced  training set (R2D2 - Scratch and Flattened)} \\
\multicolumn{3}{||c||}{(number of samples / percent of training data)} \\ \hline 
R2D2 (SF) (70k / 12\%) & 42.96 \footnotesize{$\pm 0.25$} & 41.87 \footnotesize{$\pm 0.31$}\\ 
R2D2 (SF) (5k / 1\%) & 33.18 \footnotesize{$\pm 0.34$} & 29.81 \footnotesize{$\pm 0.50$}\\ 
R2D2 (SF) (.5k / 0.1\%) & 7.67 \footnotesize{$\pm 0.78$} & 31.00 \footnotesize{$\pm 0.80$}\\
\hline 
\end{tabular}
\caption{\label{tab:fs-wikibio}Results on the WikiBio task. State-of-the-art~\citep{liu2018table} is notated by *.}
\end{table*}

\begin{table*}[h]
\centering
\renewcommand\tabcolsep{4pt}
 \begin{tabular}{||c | c c c c c||} 
 \hline
 Method &  BLEU-4 & NIST & METEOR & ROUGE-L & CIDEr\\ 
 \hline\hline
  PragInfo* & \bf 68.60 & \bf 8.73 & 45.25 & \bf 70.82 & 2.37 \\
  \hline  
   R2D2 & 67.70 \footnotesize{$\pm 0.64$}  & 8.68 \footnotesize{$\pm 0.10$} & \bf 45.85 \footnotesize{$\pm 0.18$} & 70.44 \footnotesize{$\pm 0.32$} & \bf 2.38 \footnotesize{$\pm 0.04$}\\
 R2D2 (Flattened) & 65.73 \footnotesize{$\pm 0.97$}  & 8.48 \footnotesize{$\pm 0.13$} & 45.29 \footnotesize{$\pm 0.24$} & 70.21 \footnotesize{$\pm 0.73$} &  2.32 \footnotesize{$\pm 0.05$}\\
  R2D2 (Scratch) & 52.25 \footnotesize{$\pm 2.45$}  & 7.31 \footnotesize{$\pm 0.16$} & 37.40 \footnotesize{$\pm 1.05$} & 60.13 \footnotesize{$\pm 1.76$} &  1.34 \footnotesize{$\pm 0.08$}\\
  R2D2 (SF) & 52.25 \footnotesize{$\pm 3.56$}  & 7.24 \footnotesize{$\pm 0.21$} & 36.66 \footnotesize{$\pm 2.10$} & 58.97 \footnotesize{$\pm 1.91$} &  1.31 \footnotesize{$\pm 0.07$}\\
 \hline
 \multicolumn{6}{||c||}{Reduced training set (R2D2) (number of samples / percent of training data)} \\ \hline 
 R2D2 (4k / 10\%) &  67.01 \footnotesize{$\pm 0.65$}  & 8.65 \footnotesize{$\pm 0.06$} & 45.26 \footnotesize{$\pm 0.30$} & 69.48 \footnotesize{$\pm 0.60$} &  2.29 \footnotesize{$\pm 0.06$}\\
 R2D2 (1k / 2\%) &  63.93 \footnotesize{$\pm 1.13$}  & 8.46 \footnotesize{$\pm 0.11$} & 43.90 \footnotesize{$\pm 1.18$} & 66.95 \footnotesize{$\pm 0.76$} &  2.15 \footnotesize{$\pm 0.07$}\\
 R2D2 (.5k / 1\%) & 64.26 \footnotesize{$\pm 1.22$}  & 8.48 \footnotesize{$\pm 0.10$} & 43.48 \footnotesize{$\pm 0.85$} & 66.74 \footnotesize{$\pm 1.14$} &  2.12 \footnotesize{$\pm 0.05$}\\
 \hline
 \multicolumn{6}{||c||}{Reduced training set (R2D2 Flattened) (number of samples / percent of training data)} \\ \hline 
 R2D2 (Flattened) (4k / 10\%) &  65.15 \footnotesize{$\pm 0.74$}  & 8.64 \footnotesize{$\pm 0.05$} & 44.38 \footnotesize{$\pm 0.91$} & 68.71 \footnotesize{$\pm 1.18$} &  2.22 \footnotesize{$\pm 0.08$}\\
 R2D2 (Flattened) (1k / 2\%) &  63.76 \footnotesize{$\pm 0.49$}  & 8.50 \footnotesize{$\pm 0.08$} & 43.35 \footnotesize{$\pm 0.51$} & 66.92 \footnotesize{$\pm 0.98$} &  1.63 \footnotesize{$\pm 0.84$}\\
 R2D2 (Flattened) (.5k / 1\%) & 63.02 \footnotesize{$\pm 0.42$}  & 8.55 \footnotesize{$\pm 0.02$} & 43.44 \footnotesize{$\pm 0.15$} & 66.79 \footnotesize{$\pm 0.50$} &  2.17 \footnotesize{$\pm 0.04$}\\
  \hline
 \multicolumn{6}{||c||}{Reduced training set (R2D2 Scratch) (number of samples / percent of training data)} \\ \hline 
R2D2 (Scratch) (4k / 10\%) & 52.86 \footnotesize{$\pm 1.05$} & 7.14 \footnotesize{$\pm 0.12$} & 37.69 \footnotesize{$\pm 1.41$} & 60.00 \footnotesize{$\pm 0.48$} & 1.15 \footnotesize{$\pm 0.02$}\\
R2D2 (Scratch) (1k / 2\%) & 52.72 \footnotesize{$\pm 0.21$} & 7.21 \footnotesize{$\pm 0.15$} & 34.78 \footnotesize{$\pm 0.93$} & 57.33 \footnotesize{$\pm 0.74$} & 1.26 \footnotesize{$\pm 0.03$}\\
R2D2 (Scratch) (.5k / 1\%) & 49.41 \footnotesize{$\pm 0.91$} & 7.09 \footnotesize{$\pm 0.20$} & 34.95 \footnotesize{$\pm 1.29$} & 56.40 \footnotesize{$\pm 0.90$} & 1.21 \footnotesize{$\pm 0.07$}\\
  \hline
 \multicolumn{6}{||c||}{Reduced training set (R2D2 Scratch and Flattened) (number of samples / percent of training data)} \\ \hline 
R2D2 (SF) (4k / 10\%) & 51.93 \footnotesize{$\pm 1.25$} & 7.26 \footnotesize{$\pm 0.18$} & 37.19 \footnotesize{$\pm 1.04$} & 58.88 \footnotesize{$\pm 0.79$} & 1.20 \footnotesize{$\pm 0.05$}\\
R2D2 (SF) (1k / 2\%) & 49.29 \footnotesize{$\pm 0.99$} & 6.64 \footnotesize{$\pm 0.16$} & 31.85 \footnotesize{$\pm 2.41$} & 57.08 \footnotesize{$\pm 1.28$} & 1.03 \footnotesize{$\pm 0.09$}\\
R2D2 (SF) (.5k / 1\%) & 45.29 \footnotesize{$\pm 0.37$} & 6.72 \footnotesize{$\pm 0.27$} & 33.42 \footnotesize{$\pm 1.98$} & 54.34 \footnotesize{$\pm 2.03$} & 1.07 \footnotesize{$\pm 0.06$}\\
  \hline
\end{tabular}
\caption{\label{tab:fs-e2e}Results on the E2E task. State-of-the-art~\citep{shen2019pragmatically} is notated by *.}
\end{table*}

\begin{table*}[h]
\centering
 \begin{tabular}{||c | c c||} 
 \hline
 Method &  BLEU-4 & METEOR\\ 
 \hline\hline
  DataTurner* & 52.9 & {\bf 41.9}\\
 \hline
  R2D2 & {\bf 53.77} \footnotesize{$\pm 0.86$} & 41.30 \footnotesize{$\pm 0.36$}\\ 
  R2D2 (Flattened) & 53.26 \footnotesize{$\pm 1.41$} & 40.04 \footnotesize{$\pm 0.47$}\\
  R2D2 (Scratch) &  33.25 \footnotesize{$\pm 2.16$} & 25.42 \footnotesize{$\pm 0.17$}\\  
  R2D2 (SF) &  32.62 \footnotesize{$\pm 0.07$} & 24.40 \footnotesize{$\pm 0.01$}\\  
 \hline
 \multicolumn{3}{||c||}{Reduced training set (R2D2)} \\ 
 \multicolumn{3}{||c||}{(number of samples / percent of training data)} \\ \hline 
  R2D2 (1.8k / 10\%) &  52.98 \footnotesize{$\pm 0.40$} & 40.80 \footnotesize{$\pm 0.42$}\\  
   R2D2 (360 / 2\%) &  47.58 \footnotesize{$\pm 0.41$} & 38.12 \footnotesize{$\pm 0.20$}\\  
  R2D2 (180 / 1\%) &  42.86 \footnotesize{$\pm 0.74$} & 36.23 \footnotesize{$\pm 1.09$}\\  
 \hline
     \multicolumn{3}{||c||}{Reduced training set (R2D2 Flattened)} \\ 
 \multicolumn{3}{||c||}{(number of samples / percent of training data)} \\ \hline 
  R2D2 (Flattened) (1.8k / 10\%) &  50.20 \footnotesize{$\pm 0.34$} & 39.00 \footnotesize{$\pm 0.40$}\\  
   R2D2 (Flattened) (360 / 2\%) &  46.52 \footnotesize{$\pm 0.62$} & 36.97 \footnotesize{$\pm 1.21$}\\  
  R2D2 (Flattened) (180 / 1\%) &  43.10 \footnotesize{$\pm 0.64$} & 35.19 \footnotesize{$\pm 0.41$}\\  
 \hline
     \multicolumn{3}{||c||}{Reduced training set (R2D2 Scratch)} \\ 
 \multicolumn{3}{||c||}{(number of samples / percent of training data)} \\ \hline 
R2D2 (Scratch) (1.8k / 10\%) & 30.09 \footnotesize{$\pm 1.43$} & 23.30 \footnotesize{$\pm 0.35$}\\
R2D2 (Scratch) (360 / 2\%) & 22.59 \footnotesize{$\pm 1.79$} & 18.00 \footnotesize{$\pm 1.29$}\\
R2D2 (Scratch) (180 / 1\%) & 19.93 \footnotesize{$\pm 2.25$} & 16.70 \footnotesize{$\pm 1.98$}\\
 \hline
     \multicolumn{3}{||c||}{Reduced training set (R2D2 Scratch and Flattened)} \\ 
 \multicolumn{3}{||c||}{(number of samples / percent of training data)} \\ \hline 
R2D2 (SF) (1.8k / 10\%) & 29.20 \footnotesize{$\pm 1.02$} & 22.62 \footnotesize{$\pm 0.29$}\\
R2D2 (SF) (360 / 2\%) & 19.18 \footnotesize{$\pm 1.56$} & 16.71 \footnotesize{$\pm 0.95$}\\
R2D2 (SF) (180 / 1\%) & 18.64 \footnotesize{$\pm 2.03$} & 16.23 \footnotesize{$\pm 1.75$}\\
 \hline
\end{tabular}
\caption{\label{tab:fs-webnlg}Results on the WebNLG task. State-of-the-art~\citep{harkous2020have} is notated by *.}
\end{table*}

\begin{table*}[h]
\centering
 \begin{tabular}{||c | c c ||} 
 \hline
 Method &  BLEU-4 & METEOR\\ 
 \hline\hline
  {GraphWriter*} & 14.3 \footnotesize{$\pm 1.01$}& 18.8 \footnotesize{$\pm 0.28$}\\
 \hline
  R2D2 & {\bf 17.30} \footnotesize{$\pm 0.20$} & {\bf 21.82} \footnotesize{$\pm 0.15$} \\
  R2D2 (Flattened) &  16.17 \footnotesize{$\pm 0.72$} & 20.36 \footnotesize{$\pm 0.38$} \\
  R2D2 (Scratch) &  10.60 \footnotesize{$\pm 0.22$} & 15.71 \footnotesize{$\pm 0.09$} \\
  R2D2 (SF) &  9.87 \footnotesize{$\pm 0.75$} & 14.63 \footnotesize{$\pm 0.18$} \\
  \hline 
    \multicolumn{3}{||c||}{Reduced training set (R2D2)} \\ 
 \multicolumn{3}{||c||}{(number of samples / percent of training data)} \\ \hline 
  R2D2 (5k / 13\%) & 16.83 \footnotesize{$\pm 0.18$} & 21.71 \footnotesize{$\pm 0.24$} \\
 R2D2 (1k / 2.6\%) &  14.60 \footnotesize{$\pm 0.59$} & 19.03 \footnotesize{$\pm 0.80$} \\
  R2D2 (.5k / 1\%) &  13.00 \footnotesize{$\pm 0.38$} & 17.59 \footnotesize{$\pm 0.19$} \\
 R2D2 (.1k / 0.26\%) & 10.61 \footnotesize{$\pm 1.77$} & 15.56 \footnotesize{$\pm 1.20$} \\
  \hline
    \multicolumn{3}{||c||}{Reduced training set (R2D2 Flattened)} \\ 
 \multicolumn{3}{||c||}{(number of samples / percent of training data)} \\ \hline 
  R2D2 (Flattened) (5k / 13\%) & 16.37 \footnotesize{$\pm 0.28$} & 20.68 \footnotesize{$\pm 0.33$} \\
 R2D2 (Flattened) (1k / 2.6\%) &  14.31 \footnotesize{$\pm 0.82$} & 19.15 \footnotesize{$\pm 0.97$} \\
  R2D2 (Flattened) (.5k / 1\%) &  12.58 \footnotesize{$\pm 0.14$} & 16.84 \footnotesize{$\pm 0.02$} \\
 R2D2 (Flattened) (.1k / 0.26\%) & 9.38 \footnotesize{$\pm 0.21$} & 15.00 \footnotesize{$\pm 1.27$} \\
  \hline
    \multicolumn{3}{||c||}{Reduced training set (R2D2 Scratch)} \\ 
 \multicolumn{3}{||c||}{(number of samples / percent of training data)} \\ \hline 
 R2D2 (Scratch) (5k / 13\%) & 9.45 \footnotesize{$\pm 0.31$} & 14.69 \footnotesize{$\pm 0.22$}\\
R2D2 (Scratch) (1k / 2.6\%) & 3.21 \footnotesize{$\pm 1.21$} & 8.56 \footnotesize{$\pm 0.10$}\\
R2D2 (Scratch) (.5k / 1\%) & 2.96 \footnotesize{$\pm 1.09$} & 7.01 \footnotesize{$\pm 0.39$}\\
R2D2 (Scratch) (.1k / 0.26\%) & 2.70 \footnotesize{$\pm 0.79$} & 6.77 \footnotesize{$\pm 0.76$}\\
  \hline
    \multicolumn{3}{||c||}{Reduced training set (R2D2 Scratch and Flattened)} \\ 
 \multicolumn{3}{||c||}{(number of samples / percent of training data)} \\ \hline 
R2D2 (SF) (5k / 13\%) & 7.98 \footnotesize{$\pm 0.12$} & 14.01 \footnotesize{$\pm 0.55$}\\
R2D2 (SF) (1k / 2.6\%) & 3.18 \footnotesize{$\pm 0.69$} & 8.81 \footnotesize{$\pm 0.72$}\\
R2D2 (SF) (.5k / 1\%) & 3.12 \footnotesize{$\pm 0.90$} & 8.43 \footnotesize{$\pm 0.48$}\\
R2D2 (SF) (.1k / 0.26\%) & 2.31 \footnotesize{$\pm 1.79$} & 5.70 \footnotesize{$\pm 1.25$}\\
  \hline
\end{tabular}
\caption{\label{tab:fs-agenda}Results on the AGENDA task. State-of-the-art~\citep{GraphWriter} is notated by *.}
\end{table*}

\end{document}